# Computational Techniques in Multispectral Image Processing: Application to the Syriac Galen Palimpsest

Corneliu T.C. Arsene[1], Peter E. Pormann[1], William I. Sellers[1], and Siam Bhayro[2]

[1]School of Arts, Languages and Cultures, University of Manchester, United Kingdom
[2]Department of Theology and Religion, University of Exeter, United Kingdom

Multispectral/hyperspectral image analysis has experienced much development in the last decade (Kwon et al. 2013; Wang and Chunhui 2015; Shanmugam and Srinivasa Perumal 2014; Chang 2013; Zhang and Du 2012). The application of these methods to palimpsests (Bhayro 2013; Pormann 2015; Hollaus et al. 2012) has produced significant results, enabling researchers to recover texts that would be otherwise lost under the visible overtext, by improving the contrast between the undertext and the overtext. In this paper we explore an extended number of multispectral/hyperspectral image analysis methods, consisting of supervised and unsupervised dimensionality reduction techniques (van der Maaten and Hinton 2008), on a part of the Syriac Galen Palimpsest dataset (http://www.digitalgalen.net). Of this extended set of methods, eight methods gave good results: three were supervised methods – Generalized Discriminant Analysis (GDA), Linear Discriminant Analysis (LDA), and Neighborhood Component Analysis (NCA); and the other five methods were unsupervised methods – Gaussian Process Latent Variable Model (GPLVM), Isomap, Landmark Isomap, Principal Component Analysis (PCA), and Probabilistic Principal Component Analysis (PPCA). The relative success of these methods was determined visually, using color pictures, on the basis of whether the undertext was distinguishable from the overtext, resulting in the following ranking of the methods: LDA, NCA, GDA, Isomap, Landmark Isomap, PPCA, PCA, and GPLVM. These results were compared with those obtained using the Canonical Variates Analysis (CVA) method [6,7] on the same dataset, which showed remarkably accuracy (LDA is a particular case of CVA where the objects are classified to two classes). A comparison was also made with a double thresholding and processing technique, developed as part of this project, which consists of the following: the darker overtext is carefully identified by the human operator and colored in white (threshold 1), and then the remaining undertext, which is black but not as black as the overtext was, is made even darker (threshold 2). This last technique showed some initial encouraging results, but its success depends on the human operator selecting suitable cutting values. Figure 1 shows the results and a comparison of the different computational techniques applied to page 102v-107r_B of the Syriac Galen Palimpsest data (http://www.digitalgalen.net) and for the page obtained with the ultraviolet (365 nm) illumination with green color filter (i.e. called CFUG).

Ultimately the choice of technique is based on the preferences of the person trying to read the manuscript and the precise makeup of the original document but easy access to an appropriate toolset is clearly highly desirable. Further work will consist of applying other reducing dimensionality techniques that enable the recovery of the undertext in palimpsests, as well as applying the above techniques to the rest of the Syriac Galen Palimpsest.

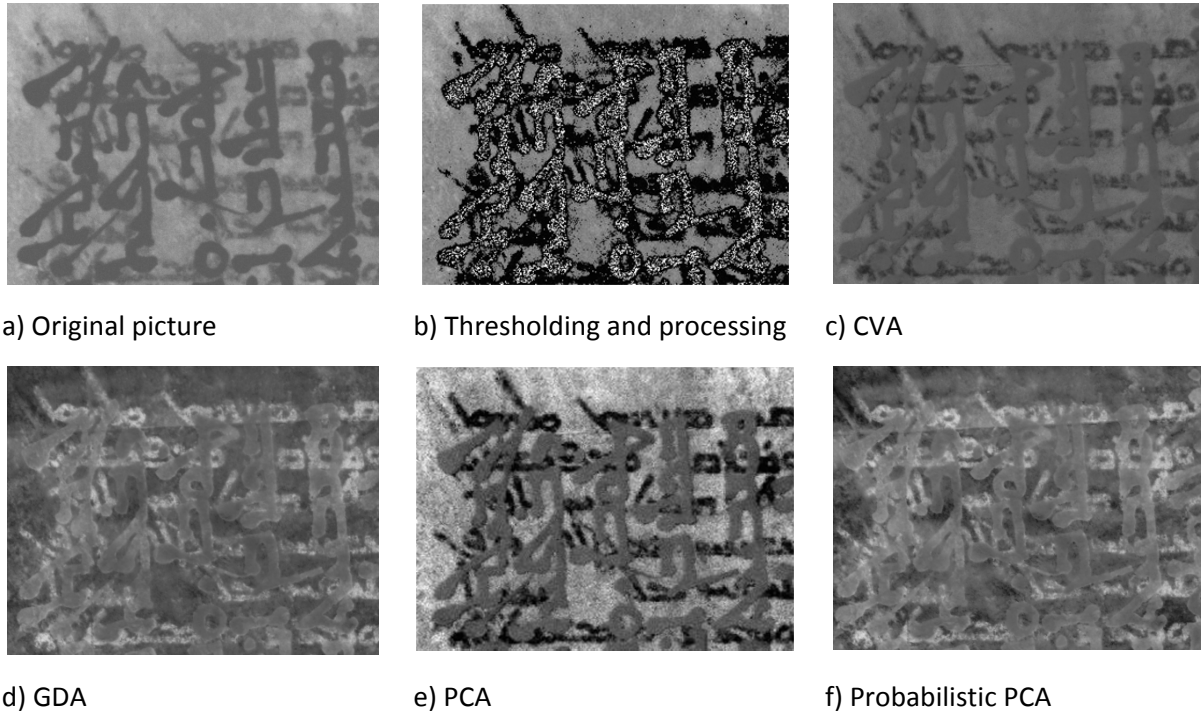

a) Original picture    b) Thresholding and processing    c) CVA

d) GDA    e) PCA    f) Probabilistic PCA

**Fig. 1:** Comparison of different computational techniques applied to the Syriac Galen Palimpsest for multispectral image processing and enhancement.

*Acknowledgment*

The authors would like to thank the Arts and Humanities Research Council, United Kingdom, for supporting this work (Research Grant AH/M005704/1 - The Syriac Galen Palimpsest: Galen's On Simple Drugs and the Recovery of Lost Texts through Sophisticated Imaging Techniques).